\documentclass[letterpaper, 10 pt, conference]{ieeeconf}  

\IEEEoverridecommandlockouts                              

\usepackage{graphics} 
\usepackage{epsfig} 
\usepackage{mathptmx} 
\usepackage{times} 
\usepackage{amsmath} 
\usepackage{ragged2e} 
\usepackage{amssymb}  
\usepackage{multirow}
\usepackage{booktabs}
\usepackage{makecell}
\usepackage{caption}
\usepackage{subfigure}
\usepackage{multicol,float}
\usepackage{algorithm}
\usepackage{cite}
\usepackage{amsmath,amssymb,amsfonts}
\usepackage{algorithmic}
\usepackage{graphicx}
\usepackage{textcomp}
\usepackage{float}
\usepackage{algorithm}
\usepackage{multirow}
\usepackage{graphicx}
\usepackage[justification=centering]{caption}
\usepackage{algorithmic}
\usepackage{booktabs}
\usepackage{xcolor}
\usepackage{bbding}
\title{\LARGE \bf
Counterfactual Graph Transformer for Traffic Flow Prediction
}

\author{Ying Yang$^{1}$, Kai Du$^{1}$, Xingyuan Dai$^{2}$, Jianwu Fang$^{1*}$
\thanks{$^{1}$Y. Yang, K. Du, and J. Fang are with the College of Transportation Engineering, Chang'an University, Xi'an, China
        {\tt\small fangjianwu@chd.edu.cn.}}%
        \thanks{$^{2}$X. Dai is with the Institute of Automation, Chinese Academy of Sciences, Beijing, China
        {\tt\small xingyuan.dai@ia.ac.cn.}}%
}

\begin{document}

\maketitle
\thispagestyle{empty}
\pagestyle{empty}

\begin{abstract}
 Traffic flow prediction (TFP) is a fundamental problem of the Intelligent Transportation System (ITS), as it models the latent spatial-temporal dependency of traffic flow for potential congestion prediction. Recent graph-based models with multiple kinds of attention mechanisms have achieved promising performance. However, existing methods for traffic flow prediction tend to inherit the bias pattern from the dataset and lack interpretability. To this end, we propose a Counterfactual Graph Transformer (CGT) model with an instance-level explainer (e.g., finding the important subgraphs) specifically designed for TFP. We design a perturbation mask generator over input sensor features at the time dimension and the graph structure on the graph transformer module to obtain spatial and temporal counterfactual explanations. By searching the optimal perturbation masks on the input data feature and graph structures, we can obtain the concise and dominant data or graph edge links for the subsequent TFP task.  After re-training the utilized graph transformer model after counterfactual perturbation, we can obtain improved and interpretable traffic flow prediction. Extensive results on three real-world public datasets show that CGT can produce reliable explanations and is promising for traffic flow prediction.
\end{abstract}

\section{Introduction}
Rapid urbanization brings heavy traffic pressure and great challenges to transportation management. As the hot spot problem of intelligent transportation systems, traffic flow prediction (TFP) is of significant importance for drivers and governments \cite{li2017diffusion}. Owning to the high nonlinearity and complexity of traffic flow, accurate and real-time TFP is challenging but with increasing attention. TFP aims to generate a series of the future amount of traffic participants based on historic observations constrained by road structure and contextual information in the traffic system. The graph \cite{yu2017spatio} is a suitable representation of road structure, and researchers often regard sensors and their connectivity as graph nodes and edges, respectively. Based on this consideration, Graph Neural Networks (GNNs) \cite{guo2020dynamic}, Recurrent Neural Networks (RNNs) \cite{laptev2017time}, and Transformer \cite{zhang2022temporal,liu2021st} are recently been widely adopted in TFP, while they easily accumulate the bias of training data because of the weak interpretability.

In highly critical scenarios, transparent and explicable models are crucial to ensure that researchers understand model action and potential bias, and to gain user trust \cite{molnar2020interpretable}. Thus, several techniques have been proposed, such as Sensitivity Analysis (SA), and Guided Backpropagation (GBP) \cite{duval2021graphsvx}. Especially, the counterfactual explainer \cite{kusner2017counterfactual} gives the chance to avoid the spurious correlation and find the causal explanation by identifying the necessary changes in the input that can alter the prediction output. Graph classification counterfactual explainer is pervasively used in social networks, molecular graphs, and transaction networks \cite{lucic2022cf,bajaj2021robust,huang2023global,feng2021should}. The similar graph-structure input of graph classification and TFP inspires us to generate the spatial-temporal counterfactual explanations of the TFP task.

\captionsetup[figure]{labelfont=bf,textfont=normalfont,singlelinecheck=off,justification=raggedright}
\begin{figure}[!t] 
\includegraphics[width=\linewidth]{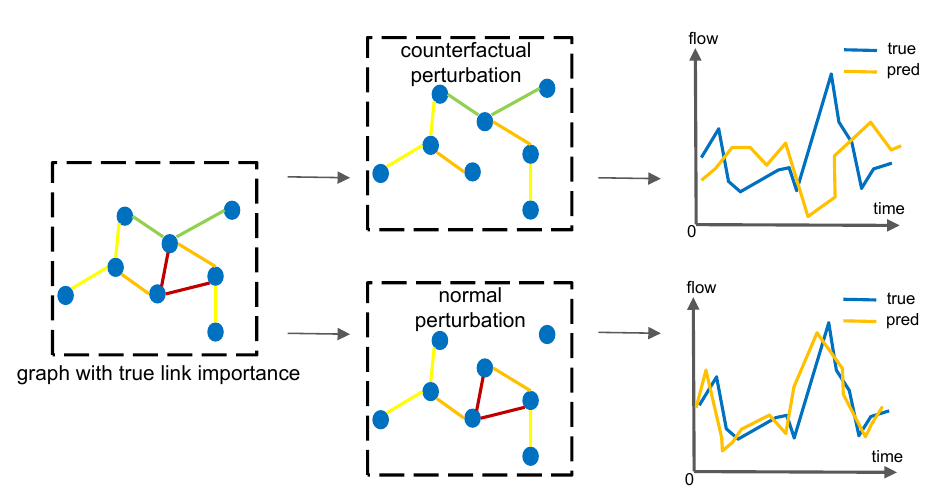}
\caption{\small{Illustration of counterfactual TFP, where the core links (i.e., red lines) of the graph will dominate the prediction.}} 
\label{fig:figure0}
\end{figure}

Instead of simply selecting a subgraph highly correlated with the prediction, counterfactual reasoning attempts to identify the smallest amount of perturbation (e.g., removing or adding edges in the graph) on the input data that significantly change the prediction \cite{baldassarre2019explainability}. Fig. \ref{fig:figure0} presents an illustration of how the counterfactual perturbation influences the performance of prediction. The red lines mean the critical links that dominate the prediction. Compared with ordinary perturbation, counterfactual perturbation aims to generate a significant impact while changing the graph structure to the least extent.
A straightforward method is to iterate over all possible perturbation masks on the input to obtain the suitable one that shows a nice counterfactual effect \cite{guo2023counterfactual}. However, it is time-consuming and cannot satisfy the validity, actionability, and sparsity properties \cite{verma2020counterfactual}.  

To address these challenges, we propose a Counterfactual Graph Transformer (CGT) model for traffic flow prediction. Given an input instance and the trained model, the counterfactual is to track perturbation causing prediction changes by deleting elements from the original input based on matrix sparsity techniques~\cite{tan2022learning}.  We propose a spatial-temporal perturbation mask on both the spatial graph and the temporally inputted data features to extract the essential perturbation for prediction. Based on this module, we design a novel counterfactual explainer to get actionable and useful explanations, which incorporates counterfactual generation and \textbf{dominant} subgraph optimization mechanisms. The dominant sub-graph is used for spatial explanations, and the search for the dominant input sensor data at different times is also explored for the temporal counterfactual explanation. Finally, we re-train the model with obtained dominant subgraphs and better TFP performance is obtained.
The \textbf{contributions} of this work are three folds:
\begin{itemize}    
\item We propose a Counterfactual Graph Transformer for the traffic prediction task. The spatial-temporal explanation can help to interpret the prediction and aid the prediction model.		
\item We design a spatial perturbation mask and a temporal instance perturbation mask and a counterfactual loss function based on the characteristic of the Graph Transformer. Our approach can effectively perturb the spatial-temporal graph structure and achieves a balance between perturbation intensity and performance degradation.	
\item We conduct extensive experiments on three real-world datasets to validate the effectiveness of the proposed method. Results show that our model reliably produces counterfactual explanations. Moreover, the explanations are used to optimize the TFP model.
\end{itemize} 
	
\section{Related Work}
\subsection{Traffic Flow Prediction}
 Currently, traffic flow prediction has been well established with the earliest research dating back to the 1970s \cite{drew1968traffic}. The categories can be divided into classical statistical models and machine learning models.  Classical statistical models are based on time series, including Historical Average (HA), Auto-Regressive Integrated Moving Average (ARIMA), and  Vector Auto-Regression (VAR) . Machine learning models, such as SVM \cite{jeong2013supervised}, KNN \cite{van2012short}, and random forests \cite{liu2017prediction}  exhibit good nonlinear fitting ability. Due to the remarkable performance of deep learning in the prediction of time series, researchers apply models, such as GNN \cite{zhao2019t,bai2021a3t},  to capture the spatial dependence of traffic flow, and the models, such as RNN \cite{li2023dynamic}, to capture the temporal dependence of traffic flow. Recently, the Transformer models \cite{yan2021learning} show a powerful ability to capture spatial-temporal information. The Spatial-Temporal Transformer Network (STTN) \cite{xu2020spatial} designs a spatial-temporal transformer to dynamically model directed spatial dependencies and the long-range bidirectional temporal dependencies. Traffic Transformer \cite{cai2020traffic} designs a global encoder and a global-local encoder to extract global spatial features and local spatial features respectively. To model the time delay of spatial information propagation, PDFormer \cite{jiang2023pdformer} proposes a traffic delay-aware feature transformation module and adaptive STTN \cite{feng2022adaptive} proposes local spatial-temporal graphs that focus on 1-hop spatial neighbors. In this paper, the CGT model employs counterfactuals to efficiently improve TFP interpretability.
\subsection{Counterfactual Explanation on GNN }
 Graph Neural Networks (GNNs) have achieved great success in representation learning on graph-based data. However, the black-box nature of GNNs hinders their widespread adoption. Many existing works \cite{luo2020parameterized} aim to identify a subgraph that is highly correlated with the classification prediction result which is likely to get misleading explanations. To avoid the spurious correlation and find causal explanations which contribute significantly to the prediction, researchers have built various models to get counterfactual explanations on graphs. CFGNNExplainer \cite{lucic2022cf} applies the \textbf{learnable sparse matrix} on a graph to generate counterfactual explanations for each node. RCExplainer \cite{bajaj2021robust} generates robust counterfactual explanations on GNNs by explicitly modeling the common decision logic of GNNs. With regard to the global view, GCFExplainer \cite{huang2023global} produces counterfactual graphs by vertex-reinforced random walks on an edit map of graphs with a greedy summary. In this work, considering graph structure together with node features, we generalize the graph classification counterfactual problem to the graph regression counterfactual problem to accommodate TFP.

\section{Problem Formulation}
\textbf{Traffic Flow Prediction :}
We use $\textbf{\textit{X}}_t\;\in\; \textbf{R}^{\;N\times C}$ to denote the traffic flow at time $\textit{t}$ of $\textit{N}$ nodes in the road network, where $C$ is the dimension of the input feature. For example, $C = 2$ when the data include the flow feature and speed feature. We define the traffic flow vector of all nodes on the whole traffic network in $T$ time steps as $\textbf{\textit{X}}=(\;\textbf{\textit{X}}_1,\;\textbf{\textit{X}}_2,\: .\:.\:.\:, \;\textbf{\textit{X}}_\textit{T})\;\in \;\textbf{R}^{\; T \times N\times C}$. We set $\textbf{\textit{Y}}=(\;\textbf{\textit{X}}_{T+1},\;\textbf{\textit{X}}_{T+2},\: .\:.\:.\:, \;\textbf{\textit{X}}_\textit{2T})$ to represent the real traffic flow vector in the next $T$ time steps. $\text{\textit{G}}(\textbf{\textit{V}},\;\textbf{\textit{E}},\;\textbf{\textit{A}})$ is the physical representation of the road network, where $\textbf{\textit{V}}=\{v_1,\:.\:.\:.\:,\;v_\textit{N}\}$ is the set of ${N}$ vertices representing the locations, $\textbf{\textit{E}} \;\subseteq{\;\textbf{\textit{V}} \;\times \;\textbf{\textit{V}}} $ is the set of edges reflecting the connectivity of two vertices, $\textbf{\textit{A}}$ is the adjacency matrix defined by the distance between locations. Traffic flow prediction model $\Phi$ encodes $\textbf{\textit{X}}$ and computes $\textbf{\textit{A}}$ to predict future traffic flow $\hat{\textbf{\textit{Y}}}=(\;\hat{\textbf{\textit{Y}}}_{T+1},\;\hat{\textbf{\textit{Y}}}_{T+2},\: .\:.\:.\:, \;\hat{\textbf{\textit{Y}}}_\textit{2T})$.
\begin{equation}
	(\;\hat{\textbf{\textit{Y}}}_{T+1},\;\hat{\textbf{\textit{Y}}}_{T+2},\: .\:.\:.\:, \;\hat{\textbf{\textit{Y}}}_\textit{2T})=\Phi(\;\textbf{\textit{X}}_1,\;\textbf{\textit{X}}_2,\: .\:.\:.\:, \;\textbf{\textit{X}}_\textit{T}\;; \;\textbf{\textit{A}}) \;
	\label{eq:2}
\end{equation}
\textbf{Counterfactual Explanation on Graph:}
We explain traffic flow prediction as a graph regression prediction task. The graph regression prediction task is defined as that: for a certain dataset with ground truth $\textbf{\textit{Y}}$, we need to learn a graph representation model to make the calculation result of this model mostly close to $\textbf{\textit{Y}}$. We apply the perturbation mask to the input $(\textbf{\textit{A}}, \;\textbf{\textit{X}})$, defined as the candidate $(\bar{\textbf{\textit{A}}}, \bar{\textbf{\textit{X}}})$. The counterfactual prediction $\bar{\textbf{\textit{Y}}}$ after perturbation desires to be very different from $\hat{\textbf{\textit{Y}}}$. A candidate can be called a counterfactual if feeding it into the prediction model $\Phi$ which produces the counterfactual prediction. Each counterfactual $(\bar{\textbf{\textit{A}}}, \;\bar{\textbf{\textit{X}}})$ corresponds to a counterfactual explanation:
\begin{equation}
    \Delta_{\textbf{A}}=\textbf{\textit{A}}\;-\;\bar{\textbf{\textit{A}}}\;,\;\Delta_{\textbf{X}}=\textbf{\textit{X}}\;-\;\bar{\textbf{\textit{X}}}.
\end{equation}
An optimal counterfactual $(\bar{\textbf{\textit{A}}}^*, \;\bar{\textbf{\textit{X}}}^*)$ is the closest one to the input $(\textbf{\textit{A}},\;\textbf{\textit{X}})$, which corresponds to the optimal counterfactual explanation $(\Delta_{\textbf{A}}^*\;,\;\Delta_{\textbf{X}}^*)$. 

Generally, given the trained model $\Phi$ and desired counterfactual prediction $\bar{\textbf{\textit{Y}}}$, we aim to identify the sub-graph (i.e., $\Delta_{\textbf{A}}$) and the subset (i.e., $\Delta_{\textbf{X}}$) of input node features which makes the prediction $\hat{\textbf{\textit{Y}}}$ close to $\bar{\textbf{\textit{Y}}}$. This problem can be transformed into an optimization problem for finding the mask with the least perturbation that can change the predicted result for the most. Following \cite{verma2020counterfactual,wachter2017counterfactual}, the optimization loss function is defined as:
    \begin{equation}
		\mathcal{L}=\mathcal{L}_{\text{pred}}(\textbf{A},\;\bar{\textbf{A}},\;\textbf{X},\;\bar{\textbf{X}}|\;\Phi)+\beta \mathcal{L}_{\text{dist}}(\textbf{A},\;\bar{\textbf{A}},\;\textbf{X},\;\bar{\textbf{X}}),
	\end{equation}
 where $\mathcal{L}_{\text{pred}}$ is a prediction loss that encourages $\Phi(\textbf{\textit{A}}, \;\textbf{\textit{X}})$ to be away from $\Phi(\overline{\textbf{\textit{A}}},\; \overline{\textbf{\textit{X}}})$, $\mathcal{L}_{\text{dist}}$ is a distance loss that encourages $(\overline{\textbf{\textit{A}}},\; \overline{\textbf{\textit{X}}})$ to be close to $(\textbf{\textit{A}}, \;\textbf{\textit{X}})$, and $\beta$ is the hyperparameter that reduces attention to $\mathcal{L}_{\text{dist}}$ when the model is updated.

\section{Counterfactual Graph Transformer}
The Counterfactual Graph Transformer consists of the \emph{Graph Transformer} and Counterfactual Explainer. Counterfactual Explainer applies the \emph{Counterfactuals Generator} and the \emph{Counterfactual Optimizer} (searching the optimal perturbation mask) to perform counterfactual explanations for a Graph Transformer in this work.  
 Optimizing Counterfactual Explainer is based on the results of two branches that consist of the Graph Transformer and the Counterfactuals Generator. The explanation helps understand the model decision process and aids prediction. We describe each module in detail.

    \subsection{Graph Transformer} \label{AA}
\begin{figure}[!t] 
		\includegraphics[width=\linewidth]{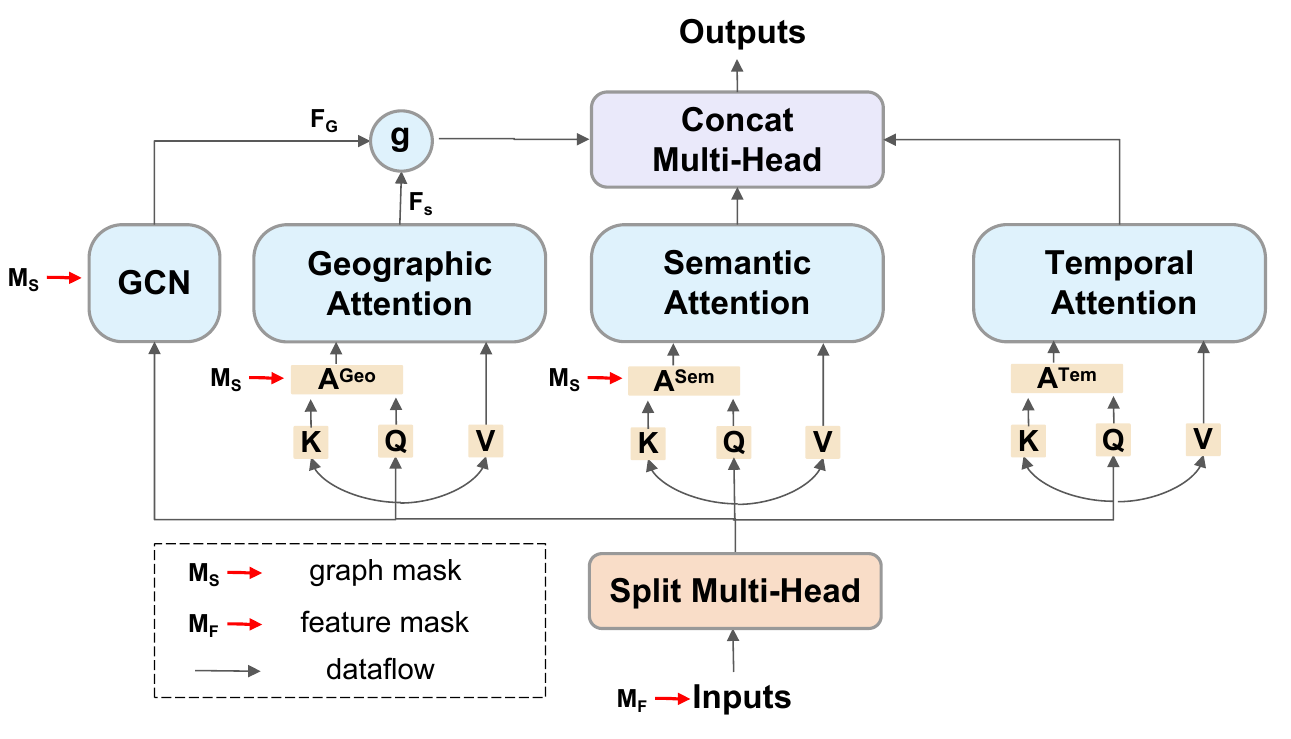}
		\caption{\small{A Graph Transformer framework (PDFormer \cite{jiang2023pdformer}) with our perturbation mask $\textbf{\textit{M}}$.}} 
		\label{fig:figure1}
	\end{figure}

	For traffic flow prediction, we introduce a well-performing state-of-the-art Graph Transformer model (e.g., PDFormer \cite{jiang2023pdformer}) as a baseline, which includes three major components: graph spatial transformer, semantic spatial transformer, and temporal transformer, as shown in Fig. \ref{fig:figure1}. The original model, on the one hand, generates the model performance baseline, and on the other hand, the trained model parameters are used as the major component of the Counterfactuals Generator. Fig. \ref{fig:figure1}  also marks the position where the perturbation mask $\textbf{\textit{M}}_S$ and $\textbf{\textit{M}}_F$ will have an impact on the model structure later.

    \textbf{Spatial-Temporal Transformer:}  Transformer can capture the complicated dependence in traffic flow prediction. In spatial dimension, combined with propagation delay, the transformer focuses its attention on proximal neighbors and semantic neighbors of nodes. In the time dimension, it can mine global dynamic time patterns\cite{xu2020spatial}. Formally, for the self-attention mechanism (\text{ATT}), we obtain the query (\textbf{\textit{Q}}), key (\textbf{\textit{K}}), and value (\textbf{\textit{V}}) matrices derived from the same input feature as:
    \begin{equation}
		\textbf{\textit{Q}}= \;\textbf{\textit{X}}\textbf{\textit{W}}_Q\;,\;\textbf{\textit{K}}= \;\textbf{\textit{X}}\textbf{\textit{W}}_K,\;\textbf{\textit{V}}=\; \textbf{\textit{X}}\textbf{\textit{W}}_V,
	\end{equation}
    where $\textbf{\textit{W}}$ is a learnable matrix of projection. Then, we define the attention weight matrix $\textbf{\textit{A}}$ to capture the dependency among all nodes. To be pointed out, the attention weight matrix can be regarded as a dynamic adjacent matrix $\textbf{\textit{A}}$ changing with input feature. We finish the self-attention operation by updating the value matrix $\textbf{\textit{V}}$ by $\textbf{\textit{A}}$:
    \begin{equation}
		\textbf{\textit{A}}=\frac{\textbf{\textit{Q}}\textbf{\textit{K}}^\top}{\sqrt{d}},\; \text{ATT}(\textbf{\textit{Q}},\;\textbf{\textit{K}},\;\textbf{\textit{V}})=\text{softmax}(\textbf{\textit{A}})\textbf{\textit{V}},
	\end{equation}
where $d$ is the dimension of the query, key, and value matrix.

\subsection{Counterfactuals Generator}\label{BB}
	 The framework of the Counterfactuals Generator is served by the trained graph transformer. The Counterfactuals Generator is fed with perturbed spatial graph structure and temporal node features. Specifically, the perturbation on the graph relates to the Laplace matrix in the GNN layer and attention weight matrix in the transformer layer, i.e., spatial perturbation act on the model structure.
	
	\textbf{Perturbation Mask:} We choose the perturbation mask method \cite{ying2019gnnexplainer} because it does not require artificially defining the size of the counterfactuals. The perturbation mask is the only learnable parameter in the counterfactuals generator. We define the perturbation matrix $\textbf{\textit{M}}_S$ and $\textbf{\textit{M}}_F$ with entries in [0,1] as:
\begin{equation}
\overline{\textbf{\textit{A}}}\;=\;\textbf{\textit{M}}_S\;\odot \;\textbf{\textit{A}}, \overline{\textbf{\textit{X}}}\;=\;\textbf{\textit{M}}_F\;\odot \;\overline{\textbf{\textit{X}}},
\end{equation}
where $\textbf{\textit{A}}$ represents all adjacent matrices in Graph Transformer. $\textbf{\textit{X}}$ represents the node features. Since the spatial adjacency matrix $\textbf{\textit{A}}$ contains $N\times N$ elements, and the input features have $T$ timestamps, we define $\textbf{\textit{M}}_S\in \textbf{R}^{N\times N}$ and $\textbf{\textit{M}}_F \in \textbf{R}^{1\times T}$. The operable range of spatial disturbance is quite large, a targeted initialization can efficiently search the counterfactuals. Considering the symmetry of the adjacency matrix, the $\textbf{\textit{M}}_S$ is set as a symmetric matrix, and only the upper triangle is set as a learnable parameter. We increase the randomness and instability of $\textbf{\textit{M}}_S$ by setting the value of $\textbf{\textit{M}}_S$ to be continuous.
	
\textbf{Counterfactual Generating Model:} In the original traffic flow prediction method \cite{jiang2023pdformer}, the adjacency matrix is embedded in the model.  As shown in Fig. \ref{fig:figure1}, We mask the graph $\textbf{\textit{A}}_{GCN}$ before the Laplacian operation of GCN and mask the weight matrix $\textbf{\textit{A}}_{Geo}$ and $\textbf{\textit{A}}_{Sem}$ of the transformer. Due to the narrow operational range of the time dimension, a very slight perturbation can achieve the counterfactual effect. Therefore, each perturbation only targets the input features in each timestamp, defined as:
	\begin{equation}
		\textbf{\textit{L}}\;=\;\textbf{\textit{I}}_N\;-\;\textbf{\textit{D}}_c^{-\frac{1}{2}} (\textbf{\textit{M}}_S\;\odot \;\textbf{\textit{A}}_{GCN})\; \textbf{\textit{D}}_c^{-\frac{1}{2}},
			\end{equation}
\begin{equation}
		\textbf{\textit{T}}_2=\textbf{\textit{L}}\textbf{\textit{X}} \textbf{\textit{W}}_{GCN},
	\end{equation}
	\begin{equation}
		{\rm ATT}\;(\textbf{\textit{Q}},\;\textbf{\textit{K}},\;\textbf{\textit{V}})\;=\;{\rm softmax}(\textbf{\textit{M}}_S\;\odot\; \textbf{\textit{A}}_{Geo})\;\textbf{\textit{V}},
	\end{equation}
 \begin{equation}
		{\rm ATT}\;(\textbf{\textit{Q}},\;\textbf{\textit{K}},\;\textbf{\textit{V}})\;=\;{\rm softmax}(\textbf{\textit{M}}_S\;\odot\; \textbf{\textit{A}}_{Sem})\;\textbf{\textit{V}},
	\end{equation}
	\begin{equation}
		\bar{\textbf{\textit{X}}}= \textbf{\textit{M}}_F\odot \textbf{\textit{X}},
	\end{equation}
    where $\textbf{\textit{L}}$ is the Laplacian matrix, $\textbf{\textit{W}}_{GCN}$ is the weights of one-layer GCN, $\textbf{\textit{I}}_N$ is an identity matrix, and $\textbf{\textit{D}}_c$ is the degree matrix of $\textbf{\textit{A}}_{GCN}$.

        \begin{figure}[tb] 
		\includegraphics[width=\linewidth]{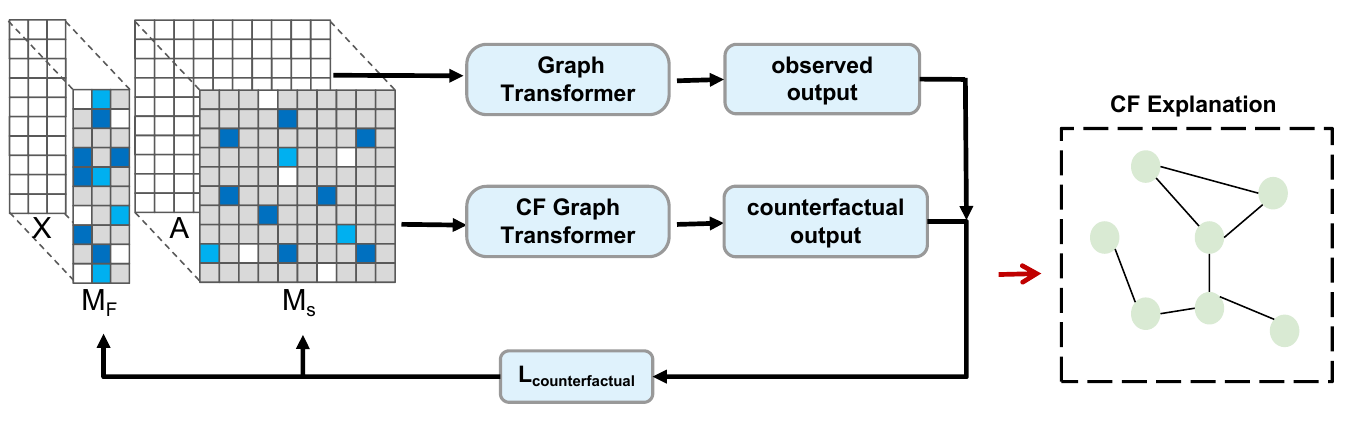}
		\caption{\small{Two branches of CGT-Explainer generate observed outcomes and counterfactual outcomes,  respectively. This process for finding the difference is called counterfactual explanation.}} 
		\label{fig:figure2}
	\end{figure} 
	
     \textbf{Geographic Gate Mechanism:} The traffic flow over a period of time can be decomposed into a stationary component determined by the road topology and a dynamic component determined by real-time traffic conditions and unexpected events \cite{xu2020spatial}.
Therefore, we apply a gate mechanism to fuse the transformer output $\textbf{\textit{T}}_1$ of Geographic Attention and the tensor $\textbf{\textit{T}}_2$ of Graph Convolution \cite{xu2020spatial} module in Fig. 2:
	\begin{equation}
		\textbf{\textit{g}}={\rm sigmoid}({\rm F_S}(\textbf{\textit{T}}_1)\;+\;{\rm F_G}(\textbf{\textit{T}}_2)),
	\end{equation}
	\begin{equation}
		\textbf{\textit{Y}}=\textbf{\textit{g}}\textbf{\textit{T}}_1+(\textbf{1}-\textbf{\textit{g}})\;\textbf{\textit{T}}_2,
	\end{equation}
where ${\rm F_S}$ and ${\rm F_G}$ are the fully-connected layers, and $ \textbf{\textit{g}}$ is the gate vector.

 \subsection{Searching $\textbf{\textit{M}}_F$ and $\textbf{\textit{M}}_S$}
    \label{AA}
	Interpretability is the main issue in graph-based TFP models. Counterfactual Optimizer aims to search for the $\textbf{\textit{M}}_F$ and $\textbf{\textit{M}}_S$ with minimal perturbation but with the largest performance change for TFP. We denote the process as a \textbf{CGT-Explainer}. The counterfactual operation extracts the features which play a crucial role in the model. Graph Transformer and Counterfactuals Generator constitute two branches of the traffic flow prediction, which respectively generate observed outcomes and counterfactual outcomes as shown in Fig. \ref{fig:figure2}. Then, after freezing all parameters inherited from the original TFP model \cite{jiang2023pdformer}, the whole counterfactual operation essentially learns a suitable perturbation mask to generate counterfactual explanations. 
    
    \begin{algorithm} 
    	\renewcommand{\algorithmicrequire}{\textbf{Require:}}
    	\caption{CGT-Explainer: Given a graph, where $\Phi(\textbf{\textit{A}})=\hat{\textbf{\textit{Y}}}$, generate the minimal perturbation $\overline{\textbf{\textit{A}}}$ with $\Phi(\overline{\textbf{\textit{A}}}) = \bar{\textbf{\textit{Y}}}$}.
    	\label{alg:1}
    	\begin{algorithmic}[1]
    		\STATE Initialization: $\hat{\textbf{\textit{M}}_S} \leftarrow \textbf{\textit{A}}, \textbf{\textit{I}}_N$   $//$ Initialization;
    		\STATE $\Phi(\textbf{\textit{A}})=\hat{\textbf{\textit{Y}}} $   $//$ Get prediction performance;
    		\STATE $\bar{\textbf{A}}^*$=[]
    		\FOR{$i=1$ to K }  
    		\STATE $\textbf{\textit{M}}_S\leftarrow threshold (\hat{\textbf{\textit{M}}_S})$
    		\STATE $\bar{\textbf{\textit{A}}}\leftarrow \textbf{\textit{M}}_S \;\odot \;\textbf{\textit{A}}$
    		\IF{$\Phi(\textbf{\textit{A}})\;\neq \;\Phi(\bar{\textbf{\textit{A}}})$} 
    		\STATE $\bar{\textbf{\textit{A}}}\leftarrow \textbf{\textit{A}} $
    		\ENDIF
    		\IF{ $\bar{\textbf{\textit{A}}} \neq \bar{\textbf{\textit{A}}}^*$ \& $D(\textbf{\textit{A}},\bar{\textbf{\textit{A}}}) \leq D(\textbf{\textit{A}},\bar{\textbf{\textit{A}}}^*)$ }
    		\STATE $\bar{\textbf{\textit{A}}}^* \leftarrow \bar{\textbf{\textit{A}}}$
    		\ENDIF 
    		\STATE $\mathcal{L}_{\hat{M_S}} \leftarrow \mathcal{L}_{\hat{M_S}}(\textbf{\textit{A}},\bar{\textbf{\textit{A}}},\Phi)$
    		\STATE $\hat{\textbf{\textit{M}}_S} \leftarrow\hat{\textbf{\textit{M}}_S}+\alpha \nabla \mathcal{L}_{\hat{M_S}}$
    		\ENDFOR
    	\end{algorithmic} 
    \end{algorithm}
	
We summarize the details of CGT-Explainer in spatial dimension (fixing $\textbf{\textit{X}}$) at \textbf{Algorithm \ref{alg:1}}. The reason why the two dimensions are not trained together is because of the huge disparity of perturbation ranges, and the best counterfactual results cannot be obtained at the same time. In Algorithm1, given the graph structure and the trained $\Phi(.)$, we first obtain its original prediction performance from $\Phi(.)$. We define a matrix $\hat{\textbf{\textit{M}}_S}$ to initially retain all edges. Later, we initialize $\hat{\textbf{\textit{M}}_S}$ according to the spacial dependency symmetry of sensor locations and the property of the original adjacency matrix. The optimization requires that the mask value be continuous, but the counterfactual explanation that we extract is to explicitly remove the useless edges. Thus, we threshold the continuous mask $\hat{\textbf{\textit{M}}_S}$ to get a binary $\textbf{\textit{M}}_S$. We apply $\textbf{\textit{M}}_S$ to sparsify the adjacent matrix $\textbf{\textit{A}}$ and get $\bar{\textbf{\textit{A}}}$. We measure the perturbation size by the difference between the two counterfactuals:
    \begin{equation}\label{e4}
    	D(\textbf{\textit{A}},\;\bar{\textbf{\textit{A}}})\; = \;sum(\;\left | \textbf{\textit{A}}\; - \;\bar{\textbf{\textit{A}}}\;\right |).
    \end{equation}	
	We keep track of the “best” counterfactual $\bar{\textbf{\textit{A}}}^*$ by sticking to the current best one when iteratively updating model parameters. We retrieve the perturbing edges and the time slice from the best counterfactual as a counterfactual explanation $(\Delta_{\textbf{A}}\;,\;\Delta_{\textbf{X}})$. For searching $\textbf{\textit{M}}_F$, we fix $\textbf{\textit{A}}$ and update the sensor input data by Eq. (11). The optimization steps of $\textbf{\textit{M}}_F$ is the same as searching $\textbf{\textit{M}}_S$.
	 	
\textbf{Loss Function Optimization:} We learn the sparse weights of $\textbf{\textit{M}}_S, \;\textbf{\textit{M}}_F$ by minimizing the following loss, adopting the Mean Squared Error (MSE) for $\mathcal{L}_{pred}$:
		\begin{equation}\label{eq:1}
            \mathcal{L}_{\text{pred}}(\textbf{\textit{A}},\;\bar{\textbf{\textit{A}}},\;\textbf{\textit{X}},\;\bar{\textbf{\textit{X}}}\;|\;\Phi)=\;-\mathcal{L}_{\text{MSE}}(\Phi(\textbf{\textit{A}},\;\textbf{\textit{X}}),\;\Phi(\bar{\textbf{\textit{A}}},\;\bar{\textbf{\textit{X}}})).
		\end{equation}
    We put a negative sign before the loss function to force the value of $\Phi(\bar{\textbf{A}})$ to keep away from $\Phi(\textbf{A})$. The definition of $\Phi(.)$ is at Equation \eqref{eq:2}. Intuitively, removing the edge of the adjacent matrix leads to terrible model performance, which means stronger perturbation makes $\mathcal{L}_{\text{pred}}$ deceptive decrease. We further propose the $\mathcal{L}_{dist}$ to limit the degree of perturbation as:
		\begin{equation}
			\mathcal{L}_{\text{dist}}(\textbf{\textit{A}},\;\bar{\textbf{\textit{A}}},\;\textbf{\textit{X}},\;\bar{\textbf{\textit{X}}}\;|\;D)=D(\textbf{\textit{A}},\;\bar{\textbf{\textit{A}}})+D(\textbf{\textit{X}},\;\bar{\textbf{\textit{X}}}),
		\end{equation}	
where $\mathcal{L}_{\text{dist}}$ represents the number of removed edges and perturbation time slices. $\mathcal{L}_{dist}$ is more feasible, and the model would preferentially decrease it, which would directly reduce $\mathcal{L}_{dist}$ to 1 and handle model convergence. Thus, we set a hyperparameter $\beta$ to balance the considerations of minimum disturbance and performance degradation when updating the model. When updating $\textbf{\textit{M}}_S$ and $\textbf{\textit{M}}_F$, we take the gradient of the loss function with respect to the continuous $\textbf{\textit{M}}_S$ and $\textbf{\textit{M}}_F$.

  \textbf{Explanation Embedding:}
    After the explainer extracts the optimal explanation $(\Delta_{\textbf{A}}^*\;\Delta_{\textbf{X}}^*)$ that dominates traffic flow prediction, we add the explanation embedding components to the Graph Transformer model to form the CGT prediction model. We increase the weight of key edges in GNN and transformer and emphasize the feature of key time slice in the input.
    
\section{Experiments}
\subsection{Datasets}
\begin{figure}[!t]
		\centering
		\includegraphics[width=7cm]{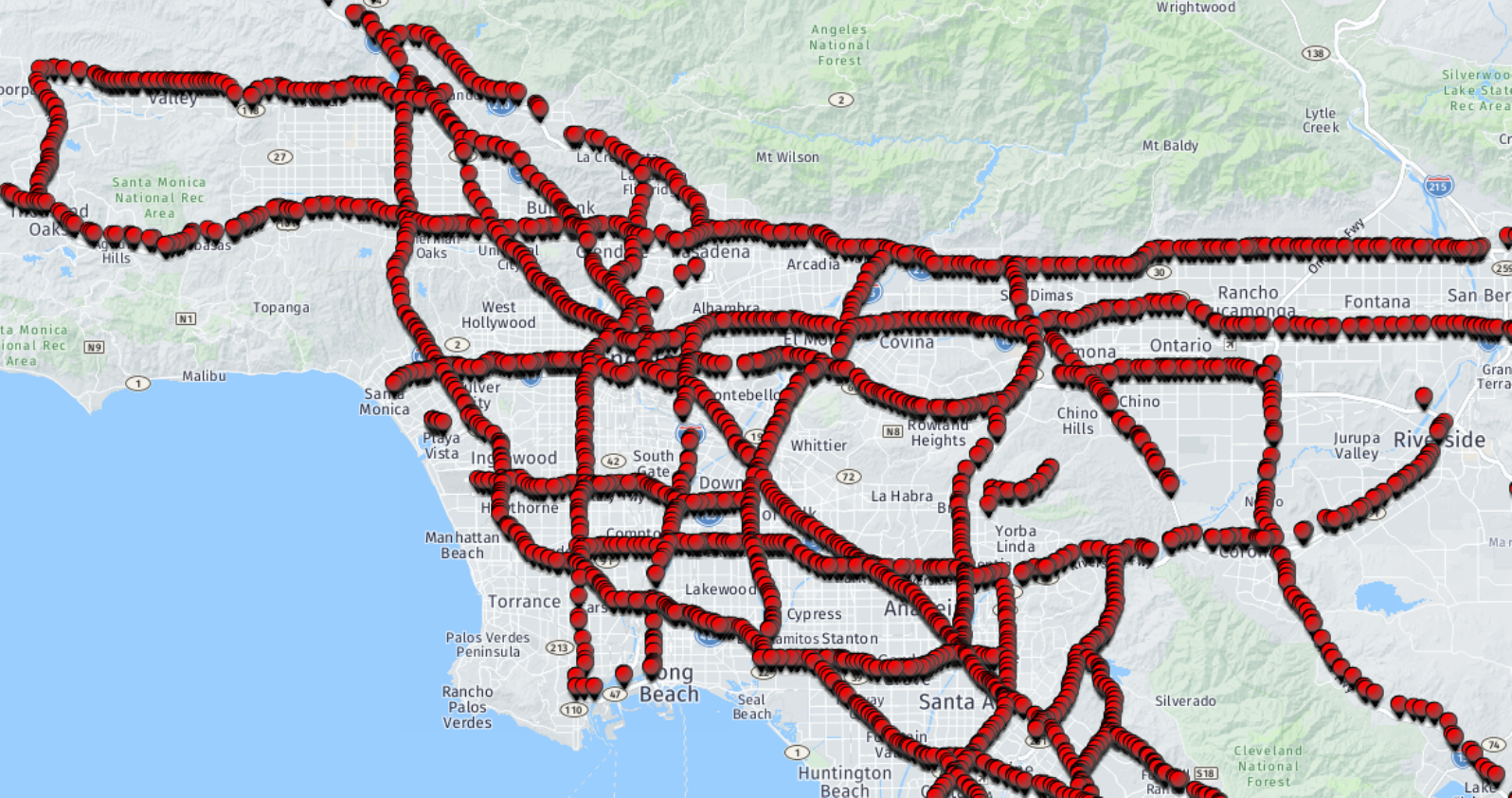}
		\caption{\small{PeMS sensor network.}}\label{fig12}
		\vspace{-1em}
	\end{figure}

	We verify the performance of CGT on three real-world traffic datasets, i.e.,  PeMS04, PeMS07M, and PeMS08 from Caltrans Performance Measurement System (PeMS) \cite{guo2019attention}. Fig. \ref{fig12} shows the location of PeMS sensors. The traffic data are uploaded every 30 seconds by sensors and then aggregated into every 5-minute interval. Details are given in Table. \ref{table:1}. Graph information about sensor locations is recorded in the datasets.
	
	\begin{table}[h!]
		\centering
		\scriptsize
		\caption{\small{Data description}.}
			  \renewcommand{\arraystretch}{1.2}
 \setlength{\tabcolsep}{0.8mm}{
		\begin{tabular}{llllll}
			\toprule
			\textbf{Datasets} & \textbf{\#Nodes} &\textbf{\#Edges} & \textbf{\#Timesteps} & \textbf{\#Time Interval} & \textbf{\#Time range}\\
			\midrule
			PeMS04 & 307 & 340 & 16992 & 5min & 01/01/2018-02/02/2018\\
			PeMS07M & 300 & 135 & 28224 & 5min & 05/01/2017-08/31/2017\\
			PeMS08 & 170 & 295 & 17856 & 5min &  07/01/2016-08/31/2016\\
			\bottomrule
		\end{tabular}}
		\label{table:1}
		\vspace{-1em}
	\end{table}

	\subsection{Baselines}
	 Considering the similarity of the input data, we design the traffic flow prediction explainer analogously to the graph counterfactual explainer. To evaluate our Explainer, we compare it against 3 explainer baselines:
	 	\begin{itemize}    
		\item RANDOM Explainer \cite{lucic2022cf}: We randomly initialize the entries of $
	\textbf{\textit{M}}_S \in [-1,1]$ and apply the same sigmoid transformation and thresholding operation as CGT as a sanity check.
		
		\item 1-hop Explainer \cite{ying2019gnnexplainer}: We search for counterfactuals from the one-hop subgraphs in the graph. For example, the one-hop subgraph of node $i$ correlates to other nodes that are reachable within a hop from node $i$. 
		
		\item GAT Explainer \cite{velivckovic2017graph}: We compute the edge’s importance by averaging attention weight across all layers, and choose the edges with the higher scores as explanations.

\end{itemize} 

In this part, we denote our method as \textbf{CGT-Explainer}. In addition, we also compare the TFP performance with four state-of-the-art methods.
	 	\begin{itemize}   
            \item MTGNN \cite{wu2020connecting}: MTGNN applies a graph learning layer to forecast time series that can automatically integrate external knowledge.
            \item STTN \cite{xu2020spatial}: STTN employs GNN and transformer to predict traffic flow, modeling long-distance temporal and spatial dependencies.
            \item ASTGNN \cite{guo2021learning}: ASTGNN employs a position embedding that contains spatial and temporal dimensions to capture Spatial heterogeneity. 
            \item PDFormer \cite{jiang2023pdformer}: PDFormer improves the ability to model the temporal delay of spatial information propagation by a traffic delay-aware feature transformation module.
\end{itemize} 
Here, we re-train the CGT after counterfactual operation again and denote it as \textbf{CGT-retrained}.

  \begin{table*}\label{Tab2}\scriptsize
		\centering
		\caption{\small{Explanation ability for our CGT-Explainer, RANDOM-Explainer, 1-hop-Explainer, and GAT-Explainer. $\downarrow$ prefers a lower value, while $\uparrow$ pursuing a higher value.}}
			  \renewcommand{\arraystretch}{1.2}
 \setlength{\tabcolsep}{1mm}{
		\begin{tabular}{c|cccc|cccc|cccc} 
			\toprule
			\multicolumn{1}{c|}{\multirow{2}{*}{Method}}& \multicolumn{4}{c}{PeMS04}& \multicolumn{4}{c}{PeMS07M}&\multicolumn{4}{c}{PeMS08}\\
			\cline{2-13} 
			&    Fidelity$\downarrow$ & e-Size$\downarrow$ & Sparsity$\uparrow$ & $\Delta$MAE $\uparrow$& Fidelity$\downarrow$ & e-Size$\downarrow$ & Sparsity$\uparrow$ & $\Delta$MAE$\uparrow$ & Fidelity$\downarrow$ & e-Size$\downarrow$ & Sparsity$\uparrow$ &$\Delta$MAE $\uparrow$\\
			\midrule
			RANDOM Explainer \cite{lucic2022cf} & \textbf{0} & 4.80  & 0.700 & \textbf{2.70}&\textbf{0}  & 13.53 & 0.980 &\textbf{1.49}& \textbf{0} &4.29 & 0.937&\textbf{2.83}\\
			1-hop Explainer \cite{ying2019gnnexplainer} &  22.28 & 3.47  & 0.950 & 0.74 & 4.31 & 4.72 & 0.931 &0.18&  6.87 & 3.74 & 0.945&1.40\\
			GAT Explainer \cite{velivckovic2017graph} & 30.28 & 2.40  & 0.965&0.39 & 6.79  & 12.40 & 0.818 &0.35& 4.68 & 2.80 & 0.965 &1.23\\
			\midrule
			CGT-Explainer-$\textbf{\textit{M}}_S^{0.5}$& 16.72 & \textbf{0.94} & \textbf{0.988} &0.84&  2.54  & \textbf{12.4}&\textbf{1.210}&0.818 & 0.001 & \textbf{2.42} & \textbf{0.954} &1.97\\
			\bottomrule
		\end{tabular}}
		\label{table:2}
	\end{table*}

	  \begin{table*}\label{Tab2}\scriptsize
		\centering
		\caption{\small{Traffic flow prediction performance on PeMS04, PeMS07M, and PeMS08 datasets.}}
		\begin{tabular}{c|ccc|ccc|ccc} 
			\toprule
			\multicolumn{1}{c}{\multirow{2}{*}{Model}}& \multicolumn{3}{|c|}{PeMS04}& \multicolumn{3}{c|}{PeMS07M}&\multicolumn{3}{c}{PeMS08}\\
			\cline{2-10} 
			&    MAE$\downarrow$ & MAPE(\%)$\downarrow$& RMSE$\downarrow$ & MAE$\downarrow$ & MAPE(\%)$\downarrow$& RMSE$\downarrow$& MAE$\downarrow$ & MAPE(\%)$\downarrow$& RMSE$\downarrow$\\
			\midrule
			MTGNN \cite{wu2020connecting} & 19.103 & 12.954  & 31.52 & 20.793&9.012  & 34.126 & 15.375 &10.084& 24.913 \\
			STTN \cite{xu2020spatial} &  19.462 & 13.627  & 31.963 & 21.379 & 9.939 & 34.572 & 15.472 &10.351&  24.953 \\
ASTGNN \cite{guo2021learning} &  18.601 & 12.630  & 31.028 & 20.616 & 8.861 & 34.017 & 14.974 &9.489&  24.710 \\
                PDFormer \cite{jiang2023pdformer} &  18.324 & 12.172  & 29.966 & 19.875 & 8.514 & 32.851 & 13.56 &9.059&  23.588 \\
				\midrule
			CGT-Explainer-$\textbf{\textit{M}}_S^{0.5}$  & 19.321& 13.584 & 30.552&21.306 & 11.295 & 33.885 & 17.530&19.609& 25.845 \\
                CGT-Explainer-$\textbf{\textit{M}}_F^{0.5}$  & 18.956 & 13.868  & 30.049 & 20.731 & 10.141  & 33.274 & 15.462 &15.552& 23.616 \\
			CGT-retrained-$\textbf{\textit{M}}_S^{0.5}$-$\textbf{\textit{M}}_F^{0.5}$ & \textbf{17.604} & \textbf{11.714} & \textbf{28.952} &\textbf{19.175}& \textbf{8.117}& \textbf{31.093}& \textbf{13.801}&\textbf{8.880} & \textbf{22.429} \\
			\bottomrule
		\end{tabular}
		\label{table:4}
	\end{table*}

	 \subsection{Experimental Settings}
	
	To be consistent with most modern methods, we split three graph-based datasets into training, validation, and test sets in a 6:2:2 ratio. In addition, we use the past hour's (every 5 minutes) data to predict the traffic flow for the next hour. In order to ensure that the characteristics of the input data have the least impact on the explanation effect, we randomly sample data to feed the two branches: the Graph Transformer (PDFormer \cite{jiang2023pdformer}) and the Counterfactuals Generator.
	
	All experiments are conducted on a machine with the NVIDIA GeForce 3070 GPU. The training epoch is set as 300, each value in the matrix $\textbf{\textit{M}}_S$ and $\textbf{\textit{M}}_F$ is initialized as 0.5 (denoted as $\textbf{\textit{M}}_S^{0.5}$ and $\textbf{\textit{M}}_F^{0.5}$), and the trade-off parameter $\beta$ is set as 0.5. We train our CGT-Explainer model using an SGD optimizer  with a learning rate $\alpha$ of 0.1. Besides, we train the CGT-re-trained using the Adam \cite{kingma2020method} optimizer with a learning rate of 0.001.
	
	Similar to the previous works \cite{ying2019gnnexplainer,lucic2022cf}, we use \textbf{Fidelity},  $\textbf{Explanation\;Size(e-Size)}$, $\textbf{Sparsity}$,  and $\Delta \textbf{MAE}$ metrics in the experiments to evaluate counterfactuals quality.
        \begin{equation}\footnotesize
		Fidelity = \sum_{h=1}^H\mathcal{L}_{\text{pred}}(\textbf{\textit{A}},\;\bar{\textbf{\textit{A}}},\;\textbf{\textit{X}},\;\bar{\textbf{\textit{X}}}\;|\;\Phi)\;/\;H,
	\end{equation} 
        \begin{equation}\footnotesize
		Explanation\;Size = \sum_{h=1}^H (S_1 +\; S_2 +\;...\:  +\;S_H)/H,
	\end{equation}  
        \begin{equation}\footnotesize
		Sparsity = \frac{K-Explanation\;Size}{K},
	\end{equation} 
        \begin{equation}\footnotesize
		\Delta MAE= MAE(\bar{\textbf{\textit{Y}}},\;{\textbf{\textit{Y}}})\;-\;MAE(\hat{\textbf{\textit{Y}}},\;\textbf{\textit{Y}}),
	\end{equation}
where $H$ is the number of counterfactual explanations, $S_i$ is the size of the $i$th counterfactual explanations, and $K$ is the number of edges in the original adjacency matrix. $\bar{\textbf{\textit{Y}}} $ is the counterfactual prediction, $\hat{\textbf{\textit{Y}}} $ is the prediction, and $\textbf{\textit{Y}}$ is the real traffic flow. A higher $\Delta$ MAE  corresponds to a better counterfactual explanation.
        
Following the previous works  \cite{jiang2023pdformer,guo2019attention}, we adopt the Mean Absolute Error (\textbf{MAE}), Mean Absolute Percentage Error (\textbf{MAPE}), and Root Mean Squared Error (\textbf{RMSE}) to evaluate traffic flow prediction.
 
\subsection{Explanation Comparison on Graph}
    
   To be comparable to other Explainers, we take the CGT-Explainer on the graph structures (i.e., on \textbf{A}$_{GCN}$, \textbf{A}$_{Geo}$, and  \textbf{A}$_{Sem}$) with perturbation mask of $\textbf{\textit{M}}_S^{0.5}$ (denoted as CGT-Explainer-$\textbf{\textit{M}}_S^{0.5}$). The explanation comparison results with other baselines are shown in Table \ref{table:2}. The bold results are the best. Relative to other baselines, CGT-Explainer generates the smallest counterfactuals (i.e., the smallest explanation size and the highest sparsity) in three datasets, which indicates that CGT-Explainer-$\textbf{\textit{M}}_S^{0.5}$ achieves the largest prediction performance degradation effect while with the least perturbation. RANDOM-Explainer has the best fidelity in all cases. However, the maximum perturbation takes longer calculation periods, and the counterfactuals get the worst explanation size. The performance of the 1-hop-Explainer is not too good in fidelity or explanation size. However, the computation time of 1-hop-Explainer is the shortest because it does not require back-propagation, which limits the search range of the counterfactuals. The fidelity value of the GAT-Explainer is relatively large, which indicates that the attention weight in GAT cannot characterize the graph edge importance.
\begin{table}[!t]
		\centering
		\footnotesize
		\caption{\small{Ablations on different settings of $\textbf{\textit{M}}_S$ and different perturbation parts by CGT-Explainer.}}
			  \renewcommand{\arraystretch}{1.2}
        \setlength{\tabcolsep}{1mm}{
		\begin{tabular}{l|cc}
			\toprule
			Variants & $\Delta$MAE$\uparrow$ & e-Size$\downarrow$ \\
			\midrule
			$\textbf{\textit{M}}_S^{0.5}$ on PDFormer \cite{jiang2023pdformer} & \textbf{0.84} & \textbf{0.94}\\
                $\textbf{\textit{M}}_S^{0.3}$ on PDFormer \cite{jiang2023pdformer} & 0.67 & 1.03\\
                \midrule
                $\textbf{\textit{M}}_S^{0.5}$ on PDFormer \cite{jiang2023pdformer} & \textbf{0.83} & \textbf{0.92} \\
                $\textbf{\textit{M}}_S^{0.5}$ on AttSTGCN \cite{guo2019attention} &  0.29 & 1.21 \\
                \midrule
                $\textbf{\textit{M}}_S^{0.5}$ on PDFormer (\textbf{A}$_{GCN}$, \textbf{A}$_{Geo}$, \textbf{A}$_{Sem}$)\cite{jiang2023pdformer}   & \textbf{0.84} & \textbf{0.95}\\
                $\textbf{\textit{M}}_S^{0.5}$ on PDFormer (\textbf{A}$_{GCN}$)\cite{jiang2023pdformer}  &  0.41 & 1.04\\
			\bottomrule
		\end{tabular}}
		\label{table:6}
		\vspace{-1em}
	\end{table}

\subsection{Ablation Study}
   In addition, we want to check the role of CGT-Explainer on different settings of $\textbf{\textit{M}}_S$ and different model parts in the following three ways: 1) the initialization of $\textbf{\textit{M}}_S$, 2) the choice of the prediction model, and 3) the usage of transformer mask. The results are shown in Table. \ref{table:6}, the bold results are the best. From this, we can draw the following conclusions.
    \begin{itemize}    
	\item Setting the initial values in $\textbf{\textit{M}}_S$ as 0.5 (denoted as $\textbf{\textit{M}}_S^{0.5}$) can bring a 0.17 gain than $\textbf{\textit{M}}_S^{0.3}$ on $\Delta$MAE and a 0.09 performance degradation than $\textbf{\textit{M}}_S^{0.3}$ on Explanation Size (e-Size). These results prove that $\textbf{\textit{M}}_S^{0.5}$ leads to better counterfactual effects.
	
	\item CGT-Explainer-$\textbf{\textit{M}}_S^{0.5}$ on PDFormer \cite{jiang2023pdformer} outperforms CGT-Explainer-$\textbf{\textit{M}}_S^{0.5}$ on AttSTGCN \cite{guo2019attention}. This is probably because PDFormer's excellent ability to capture spatial-temporal dependency makes the perturbation effect obvious. Differently, AttSTGCN \cite{guo2019attention} has a different graph structure compared with PDFormer \cite{jiang2023pdformer}, while we conduct the perturbation on the graph adjacent matrix similarly. 
 
	\item The perturbation on more graphs will find more core relations, i.e., that making perturbation on \textbf{A}$_{GCN}$, \textbf{A}$_{Geo}$, and \textbf{A}$_{Sem}$ improves $\Delta$MAE by 0.43 and with smaller Explanation Size than the one only with perturbation on \textbf{A}$_{GCN}$.
 
\end{itemize} 

\subsection{Analysis on Traffic Flow Prediction }
  The results of different traffic flow prediction models on PeMS04, PeMS07M, and PeMS08 are shown in Table. \ref{table:4}. We can observe that CGT-retrained has the best performance compared with other prediction models over three datasets. Besides, the performance of CGT-Explainer-$\textbf{\textit{M}}_S^{0.5}$ and CGT-Explainer-$\textbf{\textit{M}}_F^{0.5}$ is worse than that of PDFormer, which indicates that $\textbf{\textit{M}}_S$ and $\textbf{\textit{M}}_F$ achieve counterfactual effects with significant performance difference. According to the results of CGT-retrained-$\textbf{\textit{M}}_S^{0.5}$-$\textbf{\textit{M}}_F^{0.5}$ and PDFormer,  the explanation embedding brings an improvement on the MAE, MAPE, and RMSE values.  Therefore, the counterfactual thinking in this work presents interpretable and improved traffic flow prediction results.

\section{Conclusion}
In this work, we propose a Counterfactual Graph Transformer (CGT), a method for generating counterfactual explanations to improve traffic flow prediction. Specifically, we combine GCN and Transformer to capture dynamic spatial dependency in the Graph Transformer module. We optimize a CGT-Explainer with perturbations on three different spatial adjacent matrixes and sensor input features to generate explanations. After the counterfactual analysis, we also re-train the CGT model with only the dominant subgraph and sensor inputs after perturbation (CGT-retrained). Extensive experiments on three real-world datasets are conducted, and the results show that CGT-Explainer has better explanation ability and CGT-retrained generates superior traffic flow prediction performance to other state-of-the-art methods.
\bibliographystyle{IEEEtran}
\bibliography{ref}

\end{document}